\documentclass[a4paper]{article}

\usepackage{INTERSPEECH2015}

\usepackage{amsmath,graphicx,amsfonts,epsfig,epstopdf,datetime,algorithm,algorithmic,multirow,xcolor,subfigure}

\ninept

\title{Learning Speech Rate in Speech Recognition}

\makeatletter
\def\name#1{\gdef\@name{#1\\}}
\makeatother \name{{\em Xiangyu Zeng$^{1,3}$, Shi Yin$^{1,4}$, Dong Wang$^{*1,2}$}}

\address{$^1$CSLT, RIIT, Tsinghua University \\
  $^2$TNList, Tsinghua University \\
  $^3$Beijing University of Posts and Telecommunications\\
  $^4$Chongqing University of Posts and Telecommunicatoins\\
  {\small \tt \{zxy,yins\}@cslt.riit.tsinghua.edu.cn, wangdong99@mails.tsinghua.edu.cn}
}

\begin{document}
\maketitle
\begin{abstract}

A significant performance reduction is often observed in speech recognition when the rate of speech (ROS)
is too low or too high. Most of present approaches to addressing the ROS variation focus on the
change of speech signals in dynamic properties caused by ROS, and accordingly modify the dynamic model, e.g., the transition probabilities
of the hidden Markov model (HMM). However, an abnormal ROS changes not only the dynamic but also the static
property of speech signals, and thus can not be compensated for purely by modifying the dynamic model.

This paper proposes
an ROS learning approach based on deep neural networks (DNN), which involves an ROS feature as the input of the
DNN model and so the spectrum distortion caused by ROS can be learned and compensated for. The experimental results
show that this approach can deliver better performance for too slow and too fast utterances, demonstrating
our conjecture that ROS impacts both the dynamic and the static property of speech. In addition,
the proposed approach can be combined with the conventional HMM transition adaptation method, offering
additional performance gains.

\end{abstract}

\section{Introduction}

The change of speech rate often causes serious performance degradation for speech recognition systems in practical usage.
Different people are used to speak in different rates, and the same people may change the speech rate
utterance by utterance, or even within a single utterance, due to various factors such as expression,
emotion, environment, etc.

It has been known that the rate of speech (ROS) impacts automatic speech recognition (ASR). A low or high
ROS often causes serious performance reduction~\cite{1,2}. Therefore ROS estimation and compensation
has been a long-term focus in the ASR community.

The methods for ROS estimation can be categorized into three classes. In the first `unit segmentation' class,
speech signals are first segmented into
speech units (words, syllables or phones), and then the ROS is estimated as the number of units per second. For example~\cite{9}
uses an ASR system to recognize and segment speech signals, and~\cite{10,13} harness neural networks to detect syllable boundaries. In the
second `relevant feature' class, ROS is estimated from some relevant acoustic features, e.g., energy envelop change~\cite{2},
rhythm~\cite{11,15}, intensity and voicing~\cite{14} and sub-band energy~\cite{16}. Compared to the unit segment approach, this
approach does not need a first-pass speech transcription and so is much more light-weighted.
The final class involves various `dynamic modeling' approaches, which is based on general speech features (MFCC or Fbank, e.g.) but
designs advanced dynamic models to detect the change of speech content. For example, the Martingale framework proposed in~\cite{3},
and the convex weighting optimization method presented in~\cite{4}.

Regarding the ROS compensation, a simple approach is to train separate models for different ROS. For example in~\cite{4}, the ROS
was categorized into three classes (low, middle and high) and models were trained for each class with data belonging to it according to
the ROS. Another approach proposed in~\cite{5} compensates for ROS by normalizing the frame rate at different ROS so that
the number of frames keeps the same for different instances of a phone at different ROS levels. Probably
the most widely-adopted ROS compensation method in ASR is to adapt the transitional probabilities of the
hidden Markov model (HMM) when decoding utterances at different ROS levels~\cite{1,10}.

Most of the above approaches assume that the major impact of an abnormal ROS is on the temporal properties
of speech signals, i.e., the duration of phones, and so can be compensated for by modifying the dynamic model,
i.e., the frame rate and the HMM transition probabilities. This paper focuses on another impact of ROS: the change on
static properties of signals, i.e., the spectrum distortion. We argue that too slow or too fast speech not only changes
the duration of pronunciations, but also distort the spectrum. This distortion may be caused by the
unusual movement of articulators particularly when dealing with co-articulations, or simply by variations in gender, emotion
or intention that are not caused but indicated by ROS. The spectrum distortion can not be addressed by modifying
the dynamic model; instead, it has been to learned by a probabilistic model.

This paper proposes to learn ROS within the deep neural network (DNN) acoustic modeling framework. By introducing
the ROS as an additional input of the DNN model, the patterns caused by ROS variance can be learned in a supervised way
and hence can be compensated for in recognition. The experimental results show that ROS indeed impacts ASR performance
in a significant way, particularly when it is low. The ROS compensation can improve performance for slow and fast speech,
while almost does not hurt performance on normal speech. Combining with the HMM transition adaptation approach, we gain further
performance improvement.

The rest of the paper is organized as follows: in Section~\ref{sec:rel} some related work is described, and in Section~\ref{sec:theory}
the DNN-based ROS compensation is presented. The experiments are described in Section~\ref{sec:exp} and the paper is
concluded in Section~\ref{sec:con}.

\section{Related work}
\label{sec:rel}

This paper is related to previous work on ROS compensation, most of which has been mentioned in the introduction.
It should be highlighted that the frame rate normalization approach proposed in~\cite{5} is similar to our method in
the sense that both change the features extraction according to the ROS. The difference is that our
method introduces the ROS feature to regularize the acoustic model learning, while the work in~\cite{5} changes
the frame step size and so is still an implicit way to adjust the dynamic model.

Our proposal is also related to the multi-class training approach~\cite{4}, i.e., train different models for
different ROS. The difference is that our method does not train multiple classes explicitly, but
leverages the DNN structure to share the parameters of models for `any' ROS. In other words, the discrete indicator variable
(`slow' or `fast') in the multi-class training is replaced by a continuous indictor variable, that is, the ROS value.
We argue that this smoothed version of multi-class training can utilize the training data in a more efficient.

Finally, this work is related to DNN adaptation. For example in~\cite{18,19}, a speaker indicator in the form of an i-vector
is involved in the model training and provides better performance. This is quite similar to our approach; the
only difference is that the i-vector is replaced by ROS in our work.

\section{DNN-based ROS compensation}
\label{sec:theory}

\subsection{Impact of ROS variance}
\label{sec:impact}
We argue that the impact of ROS variance on speech signals is two-fold. In the dynamic aspect,
change on ROS causes  change on the temporal behavior, i.e., the duration of phone instances.
Different phones are impacted differently, and vowels tend to be impacted more significantly.
In the static aspect, change on ROS leads to spectrum distortion. These two impacts have been
found in acoustic research, e.g.,~\cite{li10}.

\begin{figure}
   \centering
   \begin{minipage}{8cm}
       \centering
       \includegraphics[width=5cm]{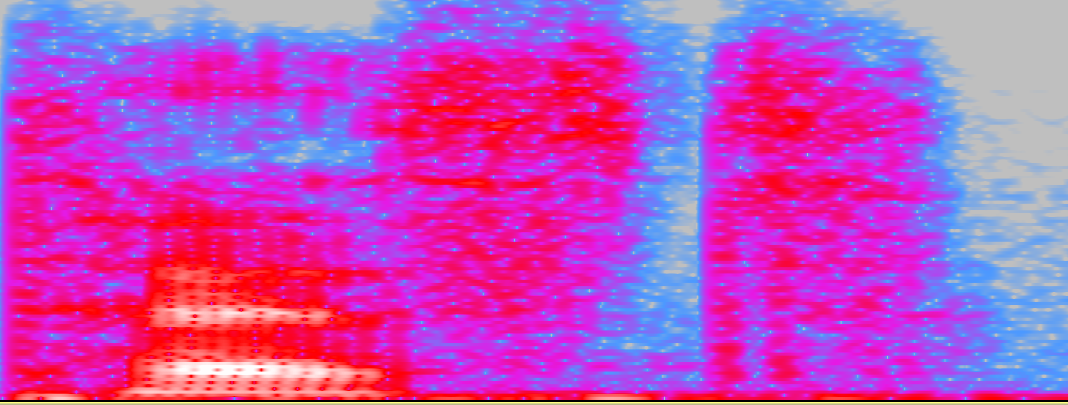}
       \caption{The spectrogram of a fast reading for word `test'.}
          \label{fig:fast}
    \end{minipage}

    \begin{minipage}{8cm}
       \centering
       \includegraphics[width=5cm]{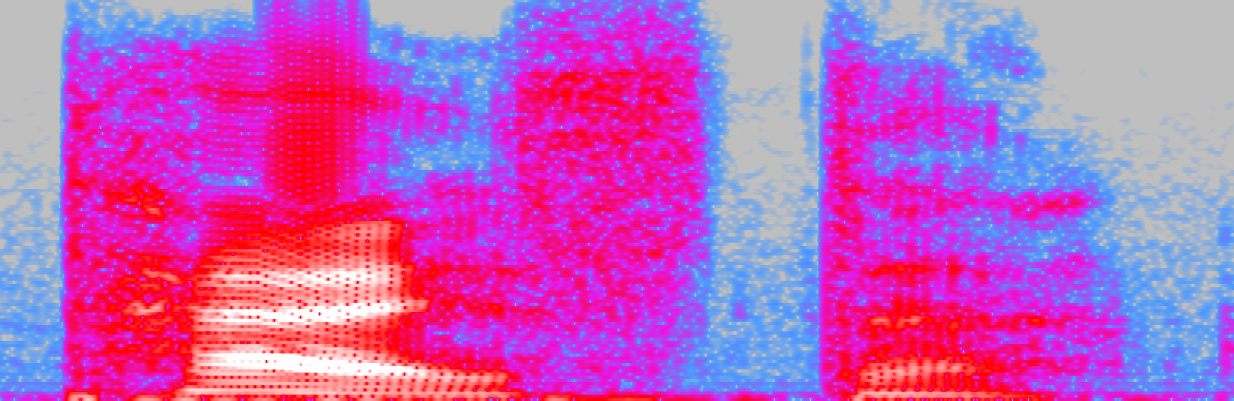}
       \caption{The spectrogram of a slow reading for word `test'.}
          \label{fig:slow}
    \end{minipage}
\end{figure}

Although the change on the dynamic property is
natural to imagine, the distortion on the static property deserves some discussion.
To have an intuition, two speech segments of the word `test' are chosen from our training database (see
Section~\ref{sec:exp}), one is clearly fast and the other is slow. The spectrograms of the two
speech signals are shown in Figure~\ref{fig:fast} and Figure~\ref{fig:slow}, respectively. Note that for
comparison, the spectrogram of the fast reading has been stretched to meet the length of the low
reading.

It can be seen that the two spectrograms are clearly different. In the slow speech, there are more formants
in the vowel part `e', and some formants shown in the consonant part `st'.  These observations demonstrate
that  ROS does cause clear distortion on speech spectrum.

\subsection{DNN-based ROS compensation}

The spectrum distortion can be compensated for by DNNs. A DNN is a special neural network that involves `deep'
structure, i.e., multiple hidden layers. Due to the deep structure, DNN possesses several advantages in
machine learning. First, it is a compact model where the units are connected and the weights are shared,
which enables it learning complex relations with limited number of parameters; second, it involves multiple
hidden layers, which makes it suitable to learn high-level features layer by layer; third, the large freedom in
the parameter space enables learning patterns in multiple conditions. Attributed to the powerful learning capability,
DNN has gained remarkable success particularly in speech recognition~\cite{deng2014,yu2015}.

Due to the advantage of DNNs in learning data in multiple conditions, it is powerful to deal with signal variations.
This capability can be leveraged to learn distortions caused by ROS, particular when the input features involves
a long-span window.
However, without an explicit indicating ROS variable, the learning could be difficult: the training needs to
discover the ROS information from the input feature and select appropriate connections to deal with various
ROS conditions. This is a `blind learning' that tends to produce moderate models for all ROS conditions.

A solution is to treat the ROS as an indicating variable and involve it in the DNN input. This simple change
turns the blind learning to an ROS-aware learning, resulting in an ROS-dependent model. This model uses
the ROS as extra information, and so can learn distortions caused by ROS.

\begin{figure}[t]
   \centering
   \includegraphics[width=\linewidth]{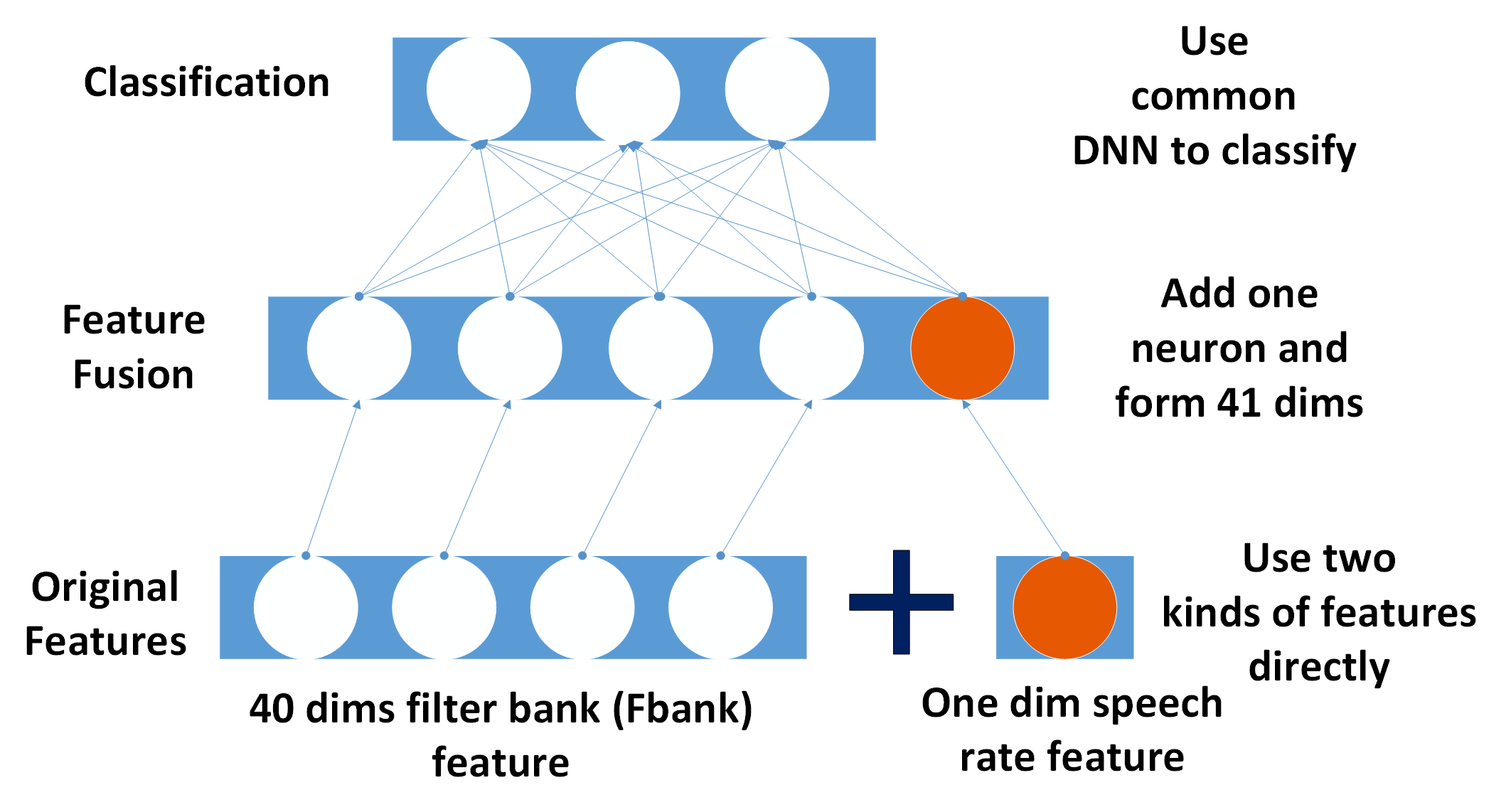}
   \caption{\it The DNN structure with ROS as an additional feature.}
   \label{fig:dnn}
\end{figure}

Figure~\ref{fig:dnn} illustrates the DNN structure we use for the ROS-aware learning. Compared to the conventional DNN
, the only difference is that the ROS is augmented to the input feature (Fbanks in our work). The training
process is identical to the one used for training standard DNNs. Note that the ROS estimation is not our focus
in this paper, and we just assume the accurate ROS has been known.

\subsection{HMM-based ROS  compensation}
\label{sec:hmm}

As mentioned, the ROS impact on the temporal property can be compensated for by modifying the dynamic model, which is
the HMM in speech recognition. The parameters that control the dynamic property of an HMM are the state transition
probabilities. It can be shown that the expectation of the duration of a phone modelled by an HMM is proportional
to the self-transition probabilities. For simplicity, assume an HMM consisting of only one state, and
the self-transition probability is $p_i$, the leaving-transition probability is accordingly $p_o=1-p_i$. The probability
that the HMM stay alive for $n$ frames is

\[
P(n) = p_i^{n-1}(1-p_i),
\]

\noindent and the expectation of the number of frames $n$ is

\[
\mathbb{E}_{P}(n) = \sum_{n=1}^{\infty} P(n) \times n = \frac{1}{p_o}
\]

\noindent Note that $ \mathbb{E}_{P}(n) \propto \frac{1}{ROS}$,  which means $ROS \propto p_o$. This relation can be
used to adjust the temporal behavior of phone HMMs so that the variance on ROS can be compensated for.

\section{Experiments}
\label{sec:exp}

\subsection{Databases}

\begin{figure}[t]
   \centering
   \includegraphics[width=6cm]{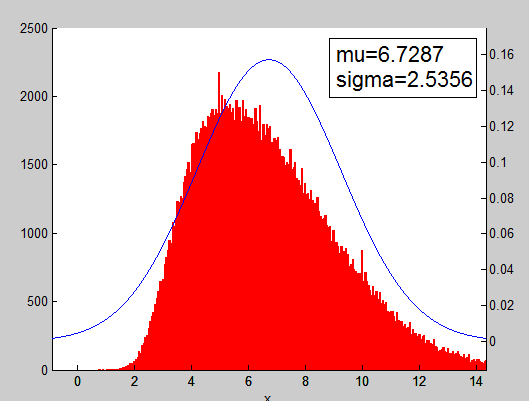}
   \caption{\it ROS distribution of the training data.}
   \label{fig:dist-train}
\end{figure}

The experiments are conducted on a Chinese spontaneous speech database provided by Tencent. The training set involves
95 hours of speech (199499 utterances), and the cross-validation (CV) set used in DNN training involves 5 hour of speech (10500 utterances).
All these utterances are collected from online applications that cover millions of people, and so the ROS variance
is more evident and realistic than most of the widely-used databases such as the wall street journal (WSJ) corpus.
Figure~\ref{fig:dist-train} shows the distribution of the ROS values of the utterances in the training dataset.
It can be seen that the distribution shows some Gaussian property as most of the ROS values concentrate in
the range of $4$-$10$ phones/second. Interestingly, the distribution exhibits a long tail in the area of
large ROS values, indicating that people tend to speak faster rather than slower.

\begin{figure}[t]
   \centering
   \includegraphics[width=6cm]{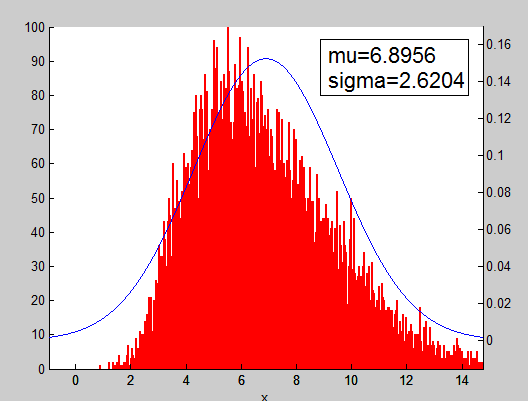}
   \caption{\it ROS distribution of the test set.}
   \label{fig:dist-test}
\end{figure}

The test set involves $6.3$ hours of speech, $10781$ utterances in total. Again, the ROS values of all the utterances
are computed and the distribution is drawn in Figure~\ref{fig:dist-test}. The distribution is similar to the one
shown in Figure~\ref{fig:dist-train}, indicating that the test data matches the training data, at least in terms
of the ROS distribution.

\begin{figure}[t]
   \centering
   \includegraphics[width=8cm]{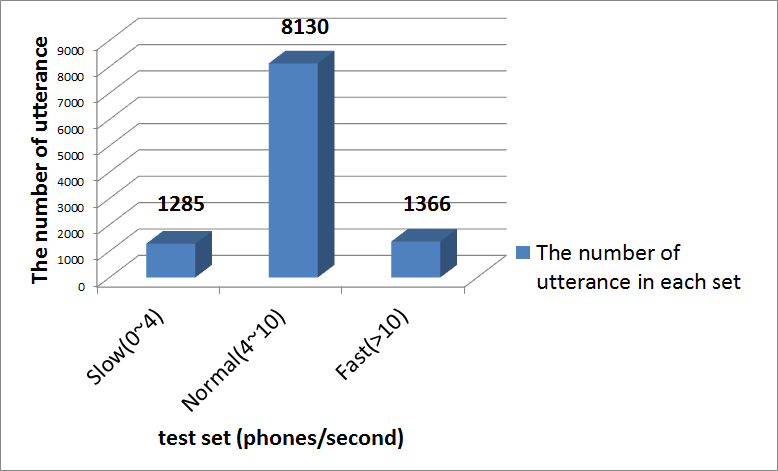}
   \caption{\it The three subsets derived from the test data.}
   \label{fig:sets}
\end{figure}

To further investigate the impact of ROS on recognition performance, the test set is divided it into three subsets: Slow ($0\sim4$ phones/s), Normal ($4\sim10$ phones/s) and Fast ($>10$ phones/s). The division is shown in Figure~\ref{fig:sets}.

\subsection{Experimental settings}

We used the Kaldi toolkit to conduct the training and evaluation, and largely followed the WSJ s5 GPU recipe. Specifically,
the first step was to establish a GMM baseline. The phone set involved $108$ Chinese initials and finals, plus a silence
phone to represent non-speech frames. The feature was 39-dimensional MFCCs, including $13$
static components plus the first- and second-order derivatives. The acoustic model was based one context-dependent
phones (tri-phones), clustered by decisions trees. After the clustering, the model consisted of $3656$
probability density functions (PDF) and the number of Gaussian components was $39995$. The GMM system was
used to produce phoneme alignments for the training data and provide the prototypes for the DNN system, including
the HMM model that describes the transition characteristics of phoneme models, and the decision tree
that describes the sharing scheme of the tri-phones.

The DNN system was then trained utilizing the phone alignments produced by the GMM system.
The $40$-dimensional Fbank feature was adopted and the cepstral mean normalization (CMN)
was employed to eliminate the effect of channel noise. In order to use dynamic information of
speech signals, the left and right $5$ frames was spliced and concatenated with the current frame.
A linear discriminant analysis (LDA) transform was used to reduce the feature dimension
to $200$. For the DNN-based ROS compensation, the ROS value was augmented to
the Fbank feature after CMN, leading to a $41$-dimensional ROS-aware feature. Again,
the left and right neighbouring frames were concatenated and the LDA was employed to
reduce the feature dimension to $200$. The LDA-transformed feature was used as the DNN input.

The DNN architecture involved $4$ hidden layers and each layer consisted of $1200$ units.
The output layer was composed of $3656$ units, equal to the total number of PDFs in
the GMM system. The training criterion was set to cross entropy, and the stochastic
gradient descendent (SGD) algorithm was employed to perform
optimization, with the mini batch size set to $256$ frames. This setting is quite
close to the GPU recipe used in Kaldi. We used a NVIDIA G760 GPU unit to perform
matrix manipulation.

\subsection{Experimental results}

\subsubsection{Baseline}

Table~\ref{tab:res} presents the baseline performance in terms of word error rate (WER). Two baselines
are reported, one is based on GMM and the other is based on DNN. It can be seen that ROS has an
significant impact on the results of both the two baselines, particularly on slow utterances. This is
consistent with the observation in Figure~\ref{fig:fast} and Figure~\ref{fig:slow}, indicating that
a slow speech tends to cause more distortion. Comparing the two baselines, it can be seen
that the DNN system outperforms the GMM system in all conditions.

\begin{table}[htb]
\caption{Baseline performance on three subsets at different ROS. }
\center
\begin{tabular}{l|cccc}
  \hline
   &\multicolumn{4}{c}{WER/\%} \\
   \hline
  Test set  &Slow &Normal &Fast & Total\\
   \hline
   ROS      &$<4$ &$4\sim10$ &$>10$ & -\\
  GMM Baseline & 57.32 & 37.44 & 40.85  &39.59 \\
  DNN Baseline & 45.71 & 28.04 & 31.22  &30.03 \\
  \hline
\end{tabular}
\footnotesize
\label{tab:res}
\end{table}

\subsubsection{DNN-based compensation}

Table~\ref{tab:dnn-comp} reports the performance with the DNN-based ROS compensation.
It can be seen that the performances on the slow and fast utterances
can be consistently improved with the ROS compensation. Interestingly, the compensation
does not impact the performance on speech at a normal speed.

\begin{table}[htb]
\caption{Performance with the DNN-based ROS compensation. }
\center
\begin{tabular}{l|cccc}
  \hline
   &\multicolumn{4}{c}{WER/\%} \\
   \hline
  Test set  &Slow &Normal &Fast & Total\\
   \hline
   ROS      &$<4$ &$4\sim10$ &$>10$ & -\\
   \hline
  DNN Baseline & 45.71 & 28.04 & 31.22 &30.03 \\
  DNN+ROS compensation & 44.92 & 28.05 & 29.54 &29.53  \\
  \hline
\end{tabular}
\footnotesize
\label{tab:dnn-comp}
\end{table}

In order to have a more clear understanding how the DNN-based ROS compensation contributes, and
compare the different behaviors of GMM and DNN systems at different ROS conditions, the test set is divided
into two subsets according to the ROS: Tst-Slow which involves the test utterances whose ROS is less than 6 phones/second,
and Tst-Fast which involves test utterances whose ROS is larger than 6 phones/second. The numbers of utterances involved in these
two sets are roughly equal. Accordingly, we divide the training data into Tr-Slow (ROS $<$ 6.3 phones/second)
and Tr-Fast (ROS $>$ 6.3 phones/second). Again, the amounts of data in the two subsets are roughly equal,
both the half of the original data volume. Finally, another training set Tr-Half is constructed
by sampling half of the utterances from the original training data. Note that the ROS distribution of Tr-Half
is the same as the original training set, and the data volume is half, equal to the volume of Tr-Slow and Tr-Fast.

\begin{table}[htb]
\caption{Performance of models trained with Tr-Half. }
\center
\begin{tabular}{l|cc}
  \hline
   &\multicolumn{2}{c}{WER\%} \\
   \hline
   Test set &Tst-Slow &Tst-Fast  \\
   \hline
   ROS &$<6$  &$>6$ \\
   \hline
  GMM baseline &45.08 &37.32 \\
  \hline
  DNN Baseline & 36.19 & 29.18  \\
  + DNN-based compensation & 35.51 & 28.70  \\
  \hline
\end{tabular}
\footnotesize
\label{tab:tr-half}
\end{table}

\begin{table}[htb]
\caption{Performance of models trained with Tr-Fast. }
\center
\begin{tabular}{l|cc}
  \hline
   &\multicolumn{2}{c}{WER\%} \\
    \hline
    Test set &Tst-Slow &Tst-Fast  \\
   \hline
   ROS  &$<6$  &$>6$ \\
   \hline
  GMM Baseline  &51.29 &36.47 \\
  \hline
  DNN Baseline & 40.36 &28.11  \\
  +DNN-based compensation & 38.42 & 27.94  \\
  \hline
\end{tabular}
\footnotesize
\label{tab:tr-fast}
\end{table}

\begin{table}[htb]
\caption{Performance of models trained with Tr-Slow.}
\center
\begin{tabular}{l|cc}
  \hline
   &\multicolumn{2}{c}{WER\%} \\
    \hline
   Test set &Tst-Slow &Tst-Fast  \\
   \hline
   ROS(phones/second) &$<6$  &$>6$ \\
   \hline
  GMM Baseline  &43.49 &42.47 \\
  \hline
  DNN Baseline & 35.35 & 36.46  \\
  +DNN-based compensation & 35.24 & 35.11  \\
  \hline
\end{tabular}
\footnotesize
\label{tab:tr-slow}
\end{table}

The three training sets (Tr-Half, Tr-Slow and Tr-Fast) are used to train the GMM and DNN systems,
and are tested on the two test sets (Tst-Slow and Tst-Fast) respectively. The results are presented
in Table~\ref{tab:tr-half}, Table~\ref{tab:tr-fast} and Table~\ref{tab:tr-slow}. The following observations
can be obtained from these results:

1) For both the GMM and DNN systems, ROS-mismatched training leads to significant performance degradation.
For example, training with Tr-Fast and testing on Tst-Slow, or vice versa. This is not surprising and indicates
that ROS has significant impact on ASR.

2) For both the GMM and DNN systems, the model trained with Tr-Half is slightly worse than the ROS-matched training, e.g., training with Tr-Fast
and testing with Tst-Fast. However it is much better than the ROS-mismatched training. This means that involving
utterances at various ROS is important to train a health ASR system.

3) From Table~\ref{tab:tr-slow}, it can be seen that training with only slow utterances seriously degrades
performance on fast utterances, but it is not the case for vice versa. This suggests that slow speech possesses
properties that are significantly different from those of normal and fast speech.

4) The DNN-based ROS compensation leads to consistent performance improvement for all the training and test conditions.
This result proved the assumption in Section~\ref{sec:theory}, that the variance on ROS
brings not only a change on duration of pronunciations, but also a change
on spectrum. The DNN-based ROS compensation presented in our paper provides a new
approach to deal with this spectrum distortion.

\subsubsection{HMM-based compensation}

It's worth to highlight that the DNN-based ROS compensation does not  modify the dynamic model (HMM), so the performance
improvement obtained in the previous experiment totally comes from the compensation for the spectrum distortion. To give
a more explicit confirmation, the conventional HMM-based compensation is implemented following the discussion in Section~\ref{sec:hmm}.
Specifically, we adjust $p_o$ to adapt the HMM to a particular ROS. In our experiment,
the self-transition probability is modified by multiplying a factor $\alpha$, and then the transition matrix is normalized
to ensure $p_o+p_1=1$. The performance is tested on the Fast and Slow subsets of the test data. For the Fast set,
$\alpha$ is set to $0.5$, and for the Slow set, $\alpha$ is set to $1.01162$.  These values are optimal on the evaluation set.

The results are presented in Table~\ref{tab:hmm}. It can be seen that the HMM-based compensation does improvement performance
on fast utterances, however for slow utterances, the contribution is not observed.
%This is quite different from the results in
%Table~\ref{tab:dnn-comp} where the performance on both the slow and fast utterances are improved by the DNN-based compensation.
This result clearly demonstrates that
the performance reduction on slow utterances (even much worse than on fast utterances, see Table~\ref{tab:res})
is not caused by temporal distortion and so can not be compensated for by adjusting HMMs.

\begin{table}[htb]
\caption{Results with the HMM-based ROS compensation. }
\center
\begin{tabular}{l|cc}
  \hline
   &\multicolumn{2}{c}{WER/\%} \\
    \hline
   Test set &Slow &Fast  \\
   \hline
   ROS       &$<4$  &$>10$ \\
   \hline
  DNN Baseline  & 45.71 & 31.22  \\
  +HMM-based compensation & 45.71 & 30.13  \\
  \hline
\end{tabular}
\footnotesize
\label{tab:hmm}
\end{table}

Finally, the DNN-based compensation and the HMM-based compensation can be
combined  together. The results are shown in Table~\ref{tab:comb}. It can be seen that the
two compensation approaches are indeed complementary and the combination provides additional performance gains.
This is a clear evidence that the ROS variance causes distortions in both the temporal and spectral domains,
and the two compensation methods address the two distortions respectively.

\begin{table}[htb]
\caption{Results with both the DNN- and HMM-based ROS compensation. }
\center
\begin{tabular}{l|cc}
  \hline
   &\multicolumn{2}{c}{WER/\%} \\
    \hline
   Test set &Slow &Fast  \\
   \hline
   ROS &$<4$  &$>10$ \\
   \hline
  DNN Baseline & 45.71 & 31.22  \\
  +DNN-based compensation & 44.92 & 29.54  \\
  +HMM-based compensation &44.76 &29.08 \\
  \hline
\end{tabular}
\footnotesize
\label{tab:comb}
\end{table}

\section{Conclusions}
\label{sec:con}

This paper presented a DNN-based compensation approach
to address the impact of ROS on speech recognition. The experimental results confirmed
our conjecture that the ROS variance causes distortions not only
in the temporal domain but also in the spectral domain. The DNN-based
ROS compensation can effectively improve performance on fast and slow
utterances, while does not impact utterances at normal speed.
When combined with the conventional HMM-based compensation, additional gains
can be achieved.

%\section{Acknowledgements}

%This research was supported by the National Science Foundation of China (NSFC) under the project No. 61371136, and the MESTDC PhD Foundation Project No. 20130002120011. It was also supported by Sinovoice and Huilan Ltd.

\newpage
\eightpt
\bibliographystyle{IEEEtran}

\bibliography{mybib}

\end{document}